# Fact-checking with Generative AI: A Systematic Cross-Topic Examination of LLMs Capacity to Detect Veracity of Political Information

Elizaveta Kuznetsova[a], Ilaria Vitulano[a], Mykola Makhortykh[b],

Martha Stolze[a], Tomas Nagy[a], Victoria Vziatysheva[b]

**Abstract**. *The purpose of this study is to assess how large language models (LLMs) can be used for fact-checking and contribute to the broader debate on the use of automated means for veracity identification. To achieve this purpose, we use AI auditing methodology that systematically evaluates performance of five LLMs—ChatGPT-4, Llama 3 (70B), Llama 3.1 (405B), Claude 3.5 Sonnet, and Google Gemini—using prompts regarding a large set of statements fact-checked by professional journalists (N=16,513). Specifically, we use topic modeling and regression analysis to investigate which factors (e.g. topic of the prompt or the LLM type) affect evaluations of true, false, and mixed statements. Our findings reveal that while ChatGPT-4 and Google Gemini achieved higher accuracy than other models, overall performance across models remains modest. Notably, the results indicate that models are better at identifying false statements, especially on sensitive topics such as COVID-19, American political controversies, and social issues, suggesting possible guardrails that may enhance accuracy on these topics. The major implication of our findings is that there are significant challenges for using LLMs for fact-checking, including significant variation in performance across different LLMs and unequal quality of outputs for specific topics which can be attributed to deficits of training data. Our research highlights the potential and limitations of LLMs in political fact-checking, suggesting potential avenues for further improvements in guardrails as well as fine-tuning.*

[a] *Weizenbaum Institute for the Networked Society, Berlin, Germany,*
*elizaveta.kuznetsova@weizenbaum-institut.de*

[b] *Institute of Communication and Media Studies, University of Bern, Bern, Switzerland*

1. Introduction

The viral launch of ChatGPT and the subsequent chatbots (e.g. Bing CoPilot) has triggered a plethora of studies looking into various applications of generative AI: education, marketing, law, and politics, to name a few. Powered by Large Language Models (LLMs), a category of generative AI models capable of generating and processing text (Naveed et al., 2023), chatbots have the potential to redefine the role of AI in the political communication landscape. Unlike earlier non-generative forms of AI, such as search engine algorithms, LLMs can not only retrieve information but also have the potential to provide interpretations of content they generate or interact with. While they can create false information or support censorship (Urman & Makhortykh, 2023), LLMs also present new opportunities for content analysis, including the assessment of information veracity, due to LLMs' ability to interpret natural language.

The use of LLMs for fact-checking is garnering increasing attention in political communication studies (Hoes et al., 2023). Since fact-checking is widely considered an important tool of democracy building (Amazeen, 2017), there has been a surge of such organizations worldwide (Graves, 2018). However, fact-checking practice is time-consuming, creating barriers for such organizations to cover a wide variety of topics and contexts. While there have been attempts to streamline fact-checking tasks, automated assessment of information veracity remains a complex challenge (Graves, 2018). Early studies have shown some optimistic results in terms of identification of misinformation and assisting academic research (Caramancion, 2023; Törnberg et al., 2023). However, significant gaps still need to be addressed. Similar to search engines and platforms mediated by non-generative AI, generative AI technologies often lack transparency (Kasneci et al., 2023). Given that LLMs' outputs are inherently probabilistic, it makes it difficult to ensure consistency of performance, potentially making them unreliable for veracity identification on sensitive topics (Makhortykh et al., 2024). Consequently, understanding how various factors, like topics of false statements, affect LLM performance in detecting information veracity is particularly important for academic research and policymakers.

This article presents the first large-scale cross-topic comparative analysis of the ability of five LLMs, ChatGPT-4, Llama 3 70B, Llama 3.1 405B, Claude 3.5 Sonnet and Google Gemini to detect the veracity of political information. Unlike earlier studies which serve as an important inspiration for the current article (Hoes et al., 2023; Kuznetsova et al., 2023; Urman & Makhortykh, 2023), we employ a large ClaimsKG-based dataset (Gangopadhyay et al., 2023) containing statements (n=16,513), evaluated by eight fact-checking organizations, and compare how different LLMs assess the veracity of these statements. Specifically, we use AI auditing methodology (Falco et al., 2021; Kuznetsova et al., 2023) to investigate how LLMs evaluate the veracity of true, false, and mixed statements depending on the topic of the claim.

We advance the existing scholarship on automated fact-checking in two ways. On the one hand, we provide an understanding of how well LLMs are at assessing the ability of LLMs to enhance fact-checking. On the other hand, we systematically test LLMs for topical bias in their judgements and suggest avenues for LLMs' improvement in the context of misinformation-related tasks.

2. Literature Review

2.1. Fact-checking in digital information environments

Fact-checking can be understood both as the routine process of verifying the truthfulness of reports or statements by comparing them with reliable evidence and as a separate genre of journalistic content (Graves, 2018). As a journalistic practice, it emerged in the 1920-1930s when the US newspapers and magazines started forming dedicated fact-checking departments (Graves & Amazeen, 2019). Yet, in the early 2000s, it rose to become a separate field of *external* fact-checking with independent groups routinely verifying different information claims, in particular in the context of politics (Graves & Amazeen, 2019). According to the 10th annual fact-checking census by the Duke Reporters' Lab there were 417 fact-checkers in 2023 in over 100 countries, a 47% increase from 2018 (Stencel et al., 2023). Currently, the International Fact-Checking Network (IFCN) at Poynter Institute features over 170 related organizations worldwide (*International Fact-Checking Network*, 2024). The American Press Institute described the aim of fact-checkers and fact-checking organizations as "re-reporting and researching the purported facts in published/recorded statements made by politicians and anyone whose words impact others' lives and livelihoods" (Jane, 2014).

Thus, fact-checking has become an important tool of debunking misinformation, albeit with varying levels of effectiveness in terms of changing people's beliefs depending on contextual and personal factors (Chan et al., 2017; Walter et al., 2020). Recent research has shown an evolving landscape of fact-checkers, ranging from public interest fact-checkers adhering to journalistic standards to those supporting the agenda of hostile actors (Montaña-Niño et al., 2024). Despite the growing popularity of this practice, manual fact-checking done by humans has its drawbacks that cannot be overcome by increasing the number of organizations involved. The process is very time-consuming, often leading to a considerable time gap between the spread of a false claim and a debunking article that comes after (Hassan et al., 2015). Another pitfall includes individual biases of human fact-checkers and varying levels of expertise, resulting in low reliability of the verification outcome (Graves & Amazeen, 2019; Nieminen & Rapeli, 2019). Additionally, the method is not easily scalable which does not allow for quick fact-checking of large amounts of data (Graves & Amazeen, 2019).

## 2.2. LLMs and automated veracity detection

With the development of computational methods, many efforts have been made to utilize them to (fully) automate fact-checking (Karadzhov et al., 2017; Nakov et al., 2021; Thorne & Vlachos, 2018; Wu et al., 2014). One of the first steps in this direction was done by Vlachos and Riedel (2014), who suggested breaking down fact-checking into several natural language processing (NLP) tasks: from detecting claims to producing verdicts and providing decision justifications. A recent survey of automated fact-checking summarized the advances in all three of the above-mentioned aspects of the fact-checking process, including an increasingly popular usage of LLMs to facilitate this process (Guo et al., 2022). LLMs have shown their ability to recognize patterns in data and, to some extent, evaluate the semantic aspects of content, which made them a promising technology for assisting or scaling fact-checking processes (Gilardi et al., 2023). The various use cases of LLMs in fact-checking can be split into four broad categories as classified by Vykopal et al. (2024): *check-worthy claim detection*, *previously fact-checked claims detection*, *evidence retrieval*, *fact verification and fake news detection*. While the first three categories of tasks have shown some promise, the automated labeling of veracity, for instance, for fake news detection, remains a challenge.

In terms of veracity predictions, current models allow for both binary labels (true/false) and more fine-grained classification options (Guo et al., 2022). Additionally, models include the category of "other" for cases when there is not enough information to definitively assess a claim (Thorne et al., 2018; Vykopal et al., 2024). However, accurate identification of a set of nuanced labels is still a hard task for computational models. Prediction accuracy for multi-labelled claims was shown to be lower compared to binary (Alhindi et al., 2018; Quelle & Bovet, 2024). Other challenges of automated fact-checking include decreased accuracy for veracity detection for content in low-resourced languages as compared to English, potential model bias and ambiguity of false claim complexity that might require consulting an expert (Nakov et al., 2021). The last issue was shown to be partially overcome by training models on domain-specific data like a set of gold standard explanations regarding public health (Kotonya & Toni, 2020), which could be useful for other topics as well.

Another concern around using LLMs for veracity detection is that the accuracy of the LLM-generated content still raises a range of concerns. For example, Makhortykh et al. (2024) tested the performance of three chatbots, Bing (currently Copilot), Bard (currently Google Gemini), and Perplexity, in answering the questions about the war in Ukraine that are often targeted and weaponized by the Kremlin. It was found that more than a quarter of the generated output did not meet the expert accuracy baseline. Furthermore, the study shows that sometimes chatbots fail to debunk misleading narratives on the war spread by the Russian government (ibid.). Research also points out a problem of hallucinations in LLMs, which can not only provide an inaccurate answer, but also invent additional details, such as fabricated eye-witness testimonies for historical topics (Makhortykh et al., 2023), thus, potentially increasing the perceived credibility of such outputs.

Another issue related to the performance of LLMs is the possibility of their performance being prone to bias. Although it was found that its degree is largely dependent on the phrasing of a prompt and context (Röttger et al., 2024), several studies attempting to measure the political bias of LLM-powered applications by administering commonly used questionnaires show a clear leaning of the generated responses towards left views (Rutinowski et al., 2023). Even though these results are limited to a very specific setting (LLM-powered application responding to the items from a questionnaire), it is still possible that political bias might affect the way in which a chatbot evaluates and fact-checks certain statements. Furthermore, even more concerning in this regard is evidence of censorship by AI chatbots. In particular, as found by Urman and Makhortykh (2023), Bard (now Google Gemini) constantly refused to answer questions about Vladimir Putin when prompted in Russian, while answering the same questions about other politicians or in other languages. Taken together, these findings indicate that the problems with chatbots fact-checking potentially misleading claims (especially political) go beyond the simple absence of information in the training data, which makes an investigation of LLM performance for such tasks even more pressing.

Previous studies highlighted ChatGPT-4's potential to label true and false statements both before and after its training data cutoff date, achieving an overall accuracy of 68.79% on fact-checked data (Hoes et al., 2023). Quelle and Bovet (2024) compared the fact-checking abilities of ChatGPT-3.5 and 4 with or without contextual information across different languages. Allowing the model to use a programmed agent enabled to make Google queries about the claim significantly increased the accuracy of predictions, while ChatGPT-4 outperformed its counterpart on average (Quelle & Bovet, 2024). Ambiguous verdicts of original labels (e.g., "half-true" and "mostly-false") as well as non-English texts determined worse results for both models (Quelle & Bovet, 2024). Studies comparing different LLMs and using information with a combination of claims for different fact-checkers are thus far limited. We have, therefore, set out to systematically examine the performance of LLMs in identifying veracity of information and ask:

RQ1. How do different LLMs perform in identifying true, false, and mixed veracity of statements?

Previous research has not only identified the lack of knowledge about the specific factors that influence model performance on fact-checking tasks (Quelle & Bovet, 2024), but has also empirically showcased that topic-level exploration can be an important avenue for veracity detection (Kuznetsova et al., 2023). In our investigation we therefore consider topics as an important factor in LLM performance and ask:

RQ2. What is the effect of the topic of claims on the accuracy of identifying true, false, and mixed statements?

## 3. Methodology

To assess how well generative AI can detect and classify veracity of information, we conducted an AI audit of five popular and widely used LLMs, ChatGPT-4, Llama 3 70B, Llama 3.1 405B, Claude 3.5 Sonnet and Google Gemini. AI auditing involves investigating the functionality and impact of algorithmic systems to understand their operation and effects. AI audits typically evaluate system performance on specific tasks, e.g., generating unsafe content (Vidgen et al., 2023) to identify errors or systematic biases. Given the increasing influence of AI-driven applications and platforms on society, AI audits are seen as essential to governance frameworks, helping to anticipate, monitor, and manage safety risks while fostering public trust in highly automated systems (Falco et al., 2021). In political communication, AI audits have become a vital tool for exploring how technology can cause systematic distortions in subject representation and subsequently affect individuals' political awareness (Rutinowski et al., 2023)

### 3.1. Data Collection

With the help of OpenAI API and Together AI, a cloud-based API supporting deployment and testing of different types of generative AI models, we prompted a structured corpus of statements published between the years 1996 and 2023 comprising known and fact-checked false claims, to examine LLMs ability to assess veracity of politics-related statements. Specifically, we collected data from ClaimsKG, a structured dataset of fact-checked claims "which facilitates structured queries about their truth values, authors, dates, journalistic reviews and other kinds of metadata" (Tchechmedjiev et al., 2019). In its updated version (Gangopadhyay et al., 2023), as of May 2024, when the data collection was conducted, ClaimsKG's dataset contained claims, fact-checked by 13 different fact-checking websites: Fullfact[1], Politifact[2], TruthOrFiction[3], Checkyourfact[4], Vishvanews[5], AFP (French)[6], AFP[7], Polygraph[8], EU factcheck[9], Factograph[10], Fatabyyano[11], Snopes[12], Africacheck[13]. Given that different fact-checkers use different methods of classification of veracity, ClaimsKG unified these ratings into four broad categories: *TRUE*, *FALSE*, *MIXTURE*, and *OTHER*. TRUE and FALSE denote the extreme cases on the veracity scale, while *"MIXTURE characterizes something on a truth scale or that holds both a degree of truth and a degree of falsehood"*. Anything outside of this spectrum was assigned to the OTHER category (Tchechmedjiev et al., 2019). From the ClaimsKG Explorer, we downloaded all statements from the years 2000 to 2023, for a total of 29,803 unique statements. Due to anticipated difficulties of automatically labelling veracity of statements with the lack of contextual information, we excluded statements consisting of a caption for videos or photos (containing words "video",

---

[1] https://fullfact.org/
[2] http://www.politifact.com/
[3] http://TruthOrFiction.com
[4] http://checkyourfact.com
[5] https://www.vishvasnews.com/
[6] https://factuel.afp.com
[7] https://factcheck.afp.com/
[8] https://www.polygraph.info/
[9] https://eufactcheck.eu/
[10] https://www.factograph.info/
[11] https://fatabyyano.net/
[12] http://www.snopes.com/
[13] https://africacheck.org/

"photo", "image", "clip"), statements in languages other than English, statements that represented direct speech (starting with personal pronouns or with the verb "Say"), statements rated as "OTHER" and those ending with a question mark. After the filtering phase, our dataset contained 16,513 statements rated as TRUE (1,724 statements), FALSE (7,989 statements) or MIXTURE (6,800 statements).

For the purpose of our study, we chose to compare the ratings provided by ClaimsKG with those from five LLMs: ChatGPT-4, Llama 3 - 70B, Llama 3.1 - 405B, Claude 3.5 Sonnet and Google Gemini. For all LLMs, we used their APIs to produce the rating for each of the 16,513 statements. To minimize variation in model performance, we set the temperature to 0.1 for all models and we designed the following prompt to have the model label our statements as TRUE, FALSE, MIXTURE or N/A:

*Please label the statement [text of the statement] from [year of publication of the statement] as TRUE, FALSE, MIXTURE, or N/A using these definitions: TRUE refers to factually accurate information without significant omissions. FALSE refers to factually inaccurate information. MIXTURE characterizes a statement that holds both a degree of truth and a degree of falsehood. If there is not enough information or context to verify the statement, label it as N/A. Present the answer as the label, followed by a comma and a short justification that is no longer than 20 words.*

We included the possibility for the models to label statements as N/A in case some of the claims would not provide enough information for the models to accurately rate them even after the statement filtering procedure outlined above.

### 3.2. Data Analysis

*Model Performance Analysis.* To assess model performance we have performed a combination of tests, employing Precision, Recall, and F1 metrics as well as intercoder reliability tests (Cohen's Kappa, Krippendorff's Alpha, as well as Brennan-Prediger tests). For the former measurement, we have considered the ClaimsKG dataset ratings as ground truth. We chose these metrics to be able to test the proportion of true positives in all the predicted positive cases (Precision) as well as the proportion of true positives in all actual positive cases (Recall). The F1 score combines the two metrics into a harmonic mean and indicates the overall model performance. In the case of intercoder reliability tests, we measured overall agreement. We have used different tests to ensure the comprehensiveness of the analysis.

*Topic Modeling and Cluster Analysis.* To systematically examine the effect of statements' topics on the ability of LLMs to determine the veracity of the information, we have performed topic modeling on the entire corpus of our baseline dataset. For this purpose, we implemented two iterations of the BERTopic algorithm (Grootendorst, 2022). The embedding process used the sentence-transformers/all-MiniLM-L6-v2 (Reimers & Gurevych, 2019) model to generate document embeddings in a 358-dimensional space. After, the UMAP dimensionality reduction technique was used. We have employed a two-iterations strategy in the clustering step. In the first iteration, HDBSCAN was used to extract topics from the data, as it has been proven to perform the best in extracting high-quality topic clusters based on dense clusters of documents (Grootendorst, 2022). For the second iteration, we used only the outlier category from the first iteration step. We assume that each statement carries semantics and, therefore, a topic could be extracted from it. This assumption was confirmed by our heuristics when we examined the claims in the outlier category. In this second step, we employed the K-means algorithm, where the number of clusters was determined by the silhouette score (Shahapure & Nicholas, 2020). As a result of the K-means algorithm, we ensured that each data point was assigned to a cluster. As a last step of the pipeline, we constructed a custom representation model for each topic cluster based on the c-TF-IDF score. Finally, we saved the BERTopic model, including the custom and embedding models, using the SafeTensors serialization method to ensure that all model aspects are preserved for study applicability. As a result, we have received 143 topic clusters: 123 from the first step and 20 from the second one.

*Topic Labeling.* Since LLMs have shown good performance in topic extraction per document (Mu et al., 2024), we used ChatGPT-4 API to provide human-readable labels to our topics. However, upon manually reviewing the list of human-readable labels, we observed considerable overlap among many topics at the keyword and semantic levels, and also, many of the topics were too small. Therefore, we conducted additional semi-supervised hierarchical clustering of the topic labels to address these two issues. For this purpose, we first embedded the topic labels using the sentence-transformers/all-MiniLM-L6-v2 model (Reimers & Gurevych, 2019) to facilitate their reorganization. Next, we clustered the labels hierarchically using the agglomerative clustering algorithm with Ward's method (Murtagh & Legendre, 2014). From the cluster analysis, using a 2.0 distance, we obtained 10 topic clusters which, for the purpose of presentation, we renamed with shorter labels. A list of the 10 topic clusters, their corresponding shortened labels, and the number of statements is presented in Table 1.

| Topic cluster name | Topic cluster shortened label | N |
|---|---|---|
| COVID-19 Pandemic: Controversies, Health Impacts, and Treatment Strategies | COVID-19 Controversies | 1,948 |
| Analysis of Economic Conditions, Labor Dynamics, and Income Inequality in the United States Across Various Administrations and Regions | U.S. Economic Analysis | 965 |
| United States Fiscal Policies, Taxation, and Economic Impact: Analysis and Influences | U.S. Fiscal Impact | 2,021 |
| US Immigration and Fiscal Policies: Impacts and Controversies | U.S. Immigration Debate | 675 |
| Contemporary Social, Legal, and Political Issues in the United States | U.S. Social Issues | 2081 |
| Societal Controversies and Debates across Varied Domains | Societal Debates | 683 |
| Contemporary Issues in United States Policy and Economics | U.S. Policy Issues | 3,404 |
| Global Political, Economic, and Social Dynamics | Global Politics | 892 |
| Controversies and Political Actions of Key Figures in North American Politics | North American Politicians' Controversies | 1,414 |
| American Politics: Governance, Policies, and Impact on Economy and Elections | American Elections and Economy | 2,430 |

*Table 1. List of clusters of topics, their shorter label and the number of statements contained.*

*Regression Analysis.* To compare ratings across LLMs and ClaimsKG we ran five logistic regressions: ChatGPT-4 and ClaimsKG, Llama 3 - 70B and ClaimsKG, Llama 3.1 - 405B and ClaimsKG, Claude 3.5 Sonnet and ClaimsKG, and Google Gemini and ClaimsKG. For each of the five comparisons, we fit a logistic regression, with the agreement between the two ratings as the dependent variable and clusters of topics as the independent variable. We additionally included an interaction term between each topic cluster and the rating from ClaimsKG. This was motivated by our interest in the effect of the topic cluster per se, and therefore we investigated the association between agreement and topic cluster for the same conditions of statement veracity. For all regressions, we excluded statements labeled by the LLMs as N/A. In addition, a few statements that were not labeled by Llama were also excluded. In all logistic regressions, we used the largest topic cluster as a reference. Hereafter, we refer to the topic clusters as topics.

An important observation is that Llama 3-70B did not perfectly follow the instructions outlined in the prompt. Some outputs, for example, started with the rating followed by an exclamation mark and not a comma as requested (e.g. "TRUE! According to Miami Dade College's 2012 report, 86 languages were spoken by its student body"). Some other statements started with a justification of the rating and presented the rating only at the end of the outcome (e.g. "Mentions of polls showing Trump beating Clinton were frequent, but not all polls showed this, and many showed the opposite. Label: MIXTURE"). In these cases, we considered Llama's labels the word "TRUE", "FALSE", "MIXTURE" and "N/A", if they appeared in capital letters in any position in the output's text. In 5 instances, Llama did not provide a label according to this definition. These statements were treated as N/A. Similarly, Gemini failed at classifying 94 statements as instructed, providing a generic error message. ChatGPT-4 reported the ratings for all statements exactly as requested.

## 4. Results

### 4.1. Evaluating Model Performance for Veracity Detection

Based on the macro F1 scores (0.36 for statements classified as true by ClaimsKG, 0.63 for mixture and 0.77 for false), ChatGPT-4 was among the best-performing LLMs across all categories of statements (*Figure 1*). Better F1 scores were achieved only for the mixture category of statements by the Claude and Gemini LLMs. Both Llama models performed worse than the other three models overall, except for the true category of statements, where Llama LLMs overperformed Gemini. As expected, between the two Llama models, the model with the higher number of parameters (Llama 3.1 405B) showed higher performance.

Overall, the performance of all LLMs turned out to be the best when evaluating false statements. While F1 scores from Llama models even in this case were borderline inacceptable (i.e. 0.58 and 0.62 macro F1 scores), the performance for the other three LLMs ranged between 0.69 to 0.77 F1 scores which is a rather high performance for a highly complex task such as veracity detection. It also exceeds the performance of LLMs for veracity detection tasks reported in earlier studies (e.g. Hoes et al., 2023).

Despite these promising results, our findings regarding the performance of LLMs for true and mixture statements turned out to be only very modest. Especially for true statements, LLMs performed particularly badly with F1 scores ranging from 0.25 to 0.36. Namely, the odds of agreement between ChatGPT-4 and ClaimsKG for a true statement (according to ClaimsKG) were lower compared to when a statement was false or mixed (See *Table 1* in *Annex*). The odds of agreement between Llama 3-70B or Llama 3.1-405 and ClaimsKG, respectively, were higher when a statement was true compared to when a statement was false or mixture (see Table 2 and 3 in the Annex).

In the case of mixture statements, the F1 scores ranged between 0.43 and 0.67. Notably, the odds of agreement between Claude 3.5 Sonnet or Google Geminia and ClaimsKG, respectively, were higher when a statement was mixture compared to when a statement was false or true (see Table 4 and 5 in the Annex).

In terms of recall, ChatGPT-4 performed better than other LLMs at identifying **false** statements (if a statement is labelled by ClaimsKG as false, ChatGPT-4 will label it as false with a probability of 80%). As for statements labelled by ClaimsKG as **true**, Llama 3 - 70B and Llama 3.1 - 405B have the highest recall (0.81 and 0.68). For statements labelled as **mixture** by ClaimsKG, Claude 3.5 Sonnet and Google Gemini perform better in terms of recall (0.7 and 0.81 respectively).

As for the precision, Llama 3 - 70B performs the best for false statements (if a statements is labelled by false by Llama 3 - 70B it will be labelled as false by ClaimsKG with 88% probability), while ChatGPT-4 performs the best for true and mixture statements. However, the precision metric as well as the F1 score is dependent on the proportion of statements submitted to the LLMs that are labelled by ClaimsKG as true/false/mixture, therefore it is less likely to generalize to other samples.

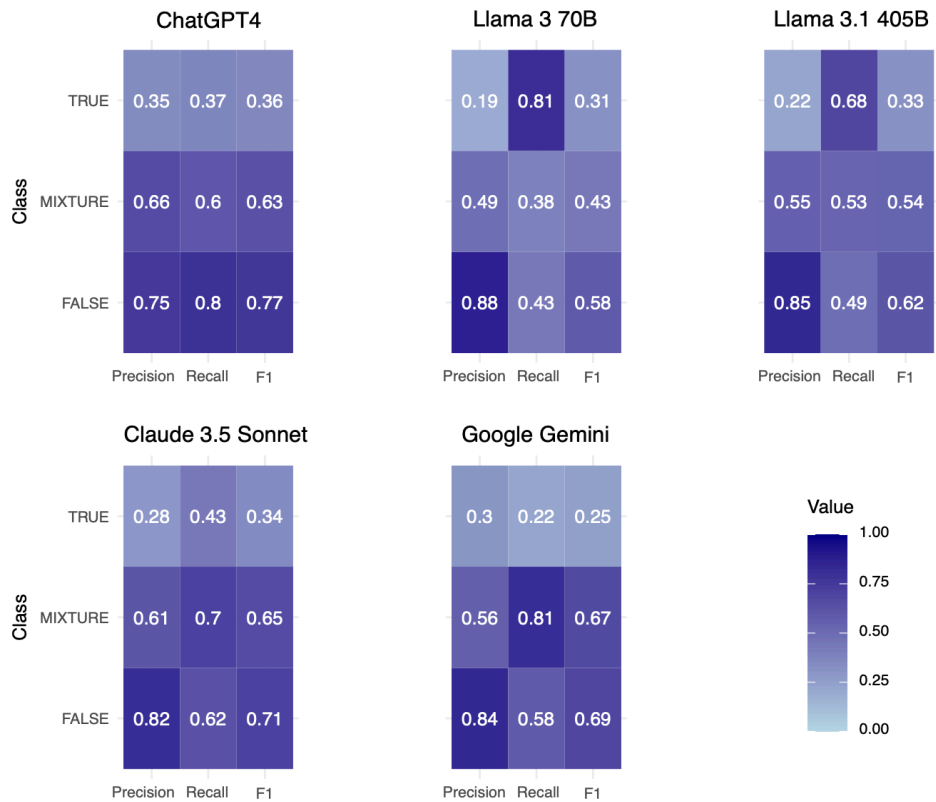

Figure *1. Precision, Recall, and F1 score comparison by veracity category between ClaimsKG and all five LLMs.*

### 4.2 Veracity Detection Performance by Topic of Claim

*Topic Effect on the Veracity Detection Performance of LLMs for **true statements***

Among the statements labeled as *true* in the ClaimsKG dataset, for the **ChatGPT-4**, **Claude 3.5 Sonnet** and **Google Gemini**, we observe no significant difference in the odds of agreement between the LLMs output and ClaimsKG by topic compared to the reference topic ("U.S. Policy Issues"). Regarding the veracity detection performance of **Llama 3 - 70B** for true statements, only two topics presented statistically significantly higher differences in the odds of agreement compared to the reference topic ("US Policy Issues"): "Global Politics" (50%) (Odds Ratio 0.504) and "U.S. Social Issues" (68%) (Odds Ratio 1.677). The only topic that presented a statistically significant difference in the odds of agreement between **Llama**

**3.1 405B** and ClaimsKG compared to the reference topic ("US Policy Issues") was also "U.S. Social Issues" (46% higher odds of agreement) (Odds Ratio 1.46) (See *Table 3* in *Annex*).

*Topic Effect on the Veracity Detection Performance of LLMs for **false statements***

As for the statements labeled as *false* by ClaimsKG, for **ChatGPT-4**, four topics presented statistically significantly higher odds of agreeing with the ClaimsKG baseline compared to the reference: "Covid-19 Controversies" (133%) , "North American Politicians' Controversies" (151%) , "Societal Debates" (54%) and "U.S. Social Issues" (65%). Interestingly, the same topics showed higher odds of agreement with ClaimsKG when using **Llama 3 - 70B**, as well as for **Llama 3.1 - 405B**, with the addition of "American Elections and Economy", as follows: "Covid19 Controversies" (148% and 102%, respectively) (2.48 and 2.02 times), "North American Politicians' Controversies" (151% and 108%) (2.51 and 2.08 times), "Societal Debates" (79% and 25%), "U.S. Social Issues" (50% and 35%), "American Elections and Economy" (38% and 30%). Similarly, for both versions of Llama LLM we tested the topic "U.S. Economic Analysis" (70% and 61%) and "U.S. Fiscal Impact" (50% and 35%) presented significantly lower odds of agreement compared to the reference. The observed substantial similarities between the two versions of the Llama LLM highlight how similarities in training data and model architecture seem to translate into similarities in fact-checking performance.

However, for **Claude 3.5 Sonnet** and **Google Gemini** the same five topics showed significantly higher odds of agreement with ClaimsKG compared to the reference topic: "Covid19 Controversies" (100% and 124% higher, respectively) (2 and 2.4 times), "North American Politicians' Controversies" (120% and 123%) (2.2 and 2.3 times), "Societal Debates" (41% and 122%), "U.S. Social Issues" (51% and 80%), "American Elections and Economy" (20% and 30%). In addition, on Google Gemini, the topic of Global Politics (48%) also showed a higher odds of agreement with the reference topic. Moreover, similar to the Llama LLMs, the topics "U.S. Economic Analysis" (44% and 67%) and "U.S. Fiscal Impact" (31% and 38%) presented significantly lower odds of agreement between Claude 3.5 Sonnet as well as Google Gemini and ClaimsKG (44% and 31% respectively), compared to the reference. These similarities are notable and arguably point to similar guardrails with regard to topics rather than purely due to similar model architecture or training data.

*Topic Effect on the Veracity Detection Performance of LLMs for **mixture statements***

Regarding the statements labelled as *mixture by ClaimsKG*, for the following topics the **ChatGPT-4** output presented significantly *lower* odds of agreeing with ClaimsKG compared to the reference topic: "American Elections and Economy" (19%), "Covid-19 Controversies" (48%), "North American Politicians' Controversies" (52%), "Societal Debates" (37%), "U.S. Immigration Debate" (25%) and "U.S. Social Issues" (23%). Concerning **Llama 3 - 70B** and **Llama 3.1 - 405B**, the output for topic "U.S. Social Issues" also presented significantly lower odds of agreement with ClaimsKG, compared to the reference topic (21% and 31% lower). Moreover, the topics "Global Politics", "U.S. Immigration Debate", and "North American Politicians' Controversies" showed significantly *higher* odds of agreement between **Llama 3-70B** and ClaimsKG compared to the reference (55%, 33% and 29%). The overlap in performance of both Llama LLMs is thus smaller regarding topics for mixture statements compared to false statements.As for **Claude 3.5 Sonnet**, only the one topic, "U.S. Fiscal Impact" showed significantly *higher* odds of agreement between the model output and ClaimsKG compared to the reference (22%), while the topics "North American Politicians' Controversies" and "Covid19 Controversies" presented *lower* odds of agreement (31% and 33% respectively). Here, both topics with lower odds of agreement overlap with ChatGPT-4 performance (see above). As for statements labeled by **Google Gemini**, four topics presented statistically *lower* odds of agreement between the LLM and ClaimsKG compared to the reference: "Covid19 Controversies" (59%), "Societal Debates" (41%), "North American Politicians' Controversies" (57%) and "U.S. Social Issues" (40%). These topics with lower odds of agreement for mixture statements on Google Gemini all overlap with the respective topics performing at lower odds on ChatGPT-4 (see above).

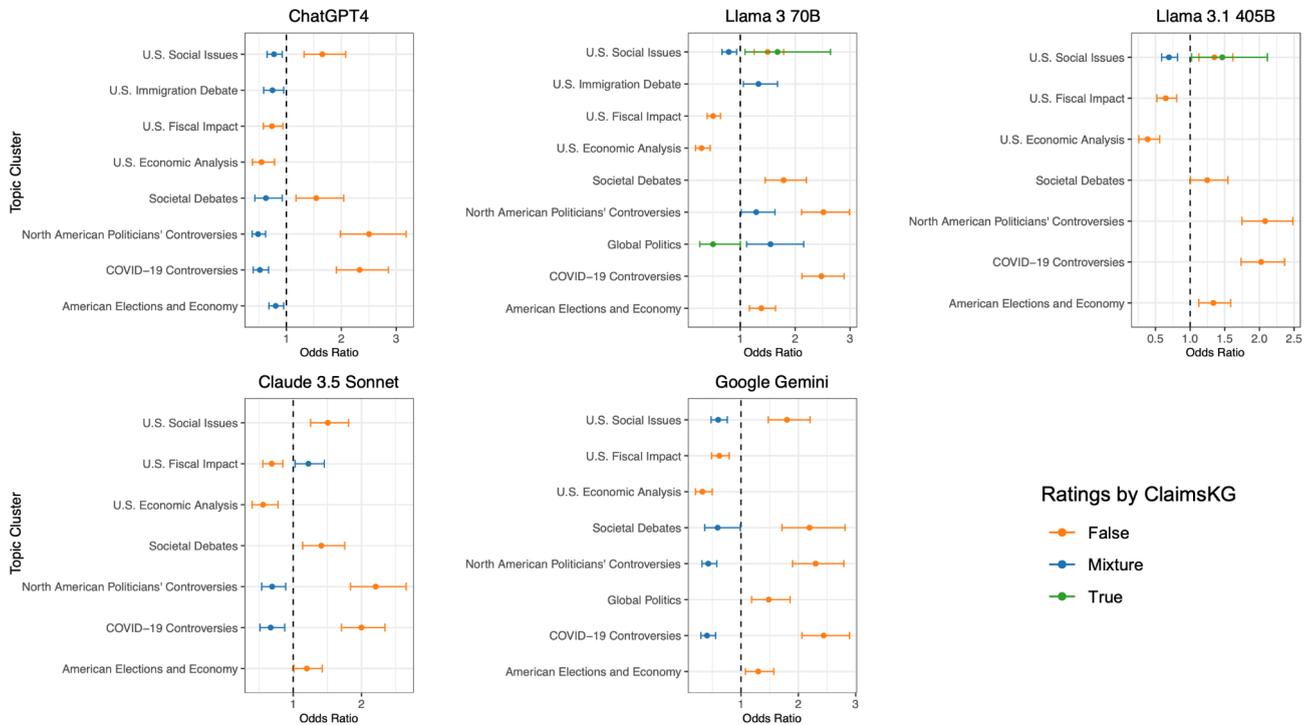

*Figure 2. Logistic regressions[14]. All five LLMs vs. ClaimsKG ratings. As a dependent variable we use the "agreement" between the models and the fact-checkers' rating from the ClaimsKG dataset, while the independent variables are the 10 topic clusters with an interaction term with the ClaimsKG rating. The reference category for the topic clusters is "U.S. Policy Issues" as it is the largest cluster. Each Odds Ratio represents the contrast between the odds of agreement between each LLM and ClaimsKG in a specific topic cluster and for a specific rating value of ClaimsKG, and the same odds in the reference topic cluster.*

**Discussion and Conclusion**

In this study, we have presented a comparative analysis of the ability of five LLMs, namely ChatGPT-4, Llama 3 - 70B, Llama 3.1 - 405B, Claude 3.5 Sonnet, Google Gemini to classify the veracity of information. To do so, we have used AI auditing methodology and systematically tested LLMs' performance in identifying true, false, and mixed categories of statements from a large dataset of fact-checked statements, ClaimsKG. We find that overall, all LLMs show a rather modest performance across all categories of statements. Generally, we find that ChatGPT-4 outperformed the other LLMs in the false and true categories of statements considering the F1 score, but underperformed for statements from the mixture category compared with Claude and Gemini.. This information is valuable to inform the choice of the LLM to use for fact-checking, in that the choice should also consider the veracity category one intends to predict with more accuracy and the proportion of true, mixture and false statements expected in the input data. This is particularly important because, in fact-checking, different errors may weigh differently under different circumstances. For instance, the classification by an LLM of a false statement as mixture in some cases may have worse implications than the classification of a true statement as mixture. However, for instance on social media platforms, the classification of a true statement as mixture can have adverse effects as it may cause a ban of legitimate opinions.

---

[14] Full regression tables are available in the Appendix:
https://osf.io/fd5p7/?view_only=bc1c39fc9a5a4b60b69485c8bc606bc0.

This performance of LLMs might potentially be further improved with targeted pre-training and fine-tuning. For example, one possible explanation for why ChatGPT-4 struggles with the true information category could be that the training data contains more fact-checked statements labeled as false and is therefore more likely to assign the corresponding label. Llama, on the other hand, is trained on different datasets and it can be the reason why it tends to over-proportionally label statements as true, even if they belong to false or mixed categories. These differences showcase that the underlying training data might have profound implications for the performance on veracity-detection tasks and should be taken into consideration when using this technology in practice. Similarly, it is important to take into consideration the model architecture (e.g. the number of parameters) which has direct implications for LLMs' performance for high complexity tasks. When taking into exam the performances of two versions of the same LLM (Llama 3 70B and Llama 3.1 405B) we notice that the latter version, trained on a vastly larger set of parameters performs better than its previous version for all categories of statements.

The above-mentioned factors can also explain our findings which highlight that there is an effect of the topic on the accuracy of veracity identification. We have observed this effect primarily within the false category of statements. All LLMs have a higher probability of labeling a statement as false when it is truly false and it belongs to the topics related to US-specific social issues, generic societal debates, American political controversies, COVID-19, and American elections, rather than to other thematic issues. Likewise, all LLMs present lower probability of labeling a statement as false when it is truly false and belongs to topics related to US fiscal issues and US economy.

Particularly noteworthy is that statements related to COVID-19 and American politics showed a higher agreement in all LLMs for the false category, potentially pointing to the possibility that all models have set up guardrails surrounding sensitive topics related to public health and politicians. While the finding could also be a result of the training data, e.g. these topics having more false statements, it is consistent with previous research on ChatGPT-4 that shows high accuracy of information produced by GPT models on health-related topics (Goodman et al., 2023). Using such guardrails could be a promising strategy for ensuring accuracy of outputs produced by LLMs. On the other hand, it is a challenging task, given that the models would need consistent adjustment to the changing socio-political context.

Our study has several limitations. Firstly, ClaimsKG dataset contains claims that are US-centric. Given that LLMs are over-proportionally trained on English-language data, the performance will most likely be different in different socio-political contexts and languages, which should be considered in future research. As LLMs can be expected to underperform in languages other than English, we are likely underreporting the performance of LLMs' fact-checking capacity. Secondly, fact-checking organizations differ in the way in which they identify statements and therefore pose a challenge for using them as ground truth. We acknowledge the need for additional filtering of fact-checked statements as not all of them provide sufficient context to be evaluated by LLMs. Lastly, while our study offers an important contribution to understanding the potential of LLMs for misinformation detection, we need to consider that it is observed for specific points in time and specific model versions. We therefore suggest that future research would benefit from reproducing these findings to keep track of models' performance.